\documentclass[sigconf]{acmart}

\AtBeginDocument{%
  }

\copyrightyear{2026}
\acmYear{2026}
\setcopyright{cc}
\setcctype{by}
\acmConference[KDD '26]{Proceedings of the 32nd ACM SIGKDD Conference on Knowledge Discovery and Data Mining V.2}{August 09--13, 2026}{Jeju Island, Republic of Korea}
\acmBooktitle{Proceedings of the 32nd ACM SIGKDD Conference on Knowledge Discovery and Data Mining V.2 (KDD '26), August 09--13, 2026, Jeju Island, Republic of Korea}
\acmDOI{10.1145/3770855.3818650}
\acmISBN{979-8-4007-2259-2/2026/08}

\usepackage{booktabs}
\usepackage{multirow}

\begin{document}

\title{Flow Learners for PDEs: Toward a Physics-to-Physics Paradigm for Scientific Computing}

\author{Yilong Dai}
\orcid{0009-0005-7459-1768}
\affiliation{%
  \institution{The University of Alabama}
  \city{Tuscaloosa}
  \state{Alabama}
  \country{USA}}
\email{ydai17@ua.edu}

\author{Shengyu Chen}
\orcid{0009-0006-9576-9841}
\affiliation{%
  \institution{University of Pittsburgh}
  \city{Pittsburgh}
  \state{Pennsylvania}
  \country{USA}}
\email{shc160@pitt.edu}

\author{Xiaowei Jia}
\orcid{0000-0001-8544-5233}
\affiliation{%
  \institution{University of Pittsburgh}
  \city{Pittsburgh}
  \state{Pennsylvania}
  \country{USA}}
\email{xiaowei@pitt.edu}

\author{Runlong Yu}
\authornote{Corresponding author.}
\orcid{0000-0003-4080-2377}
\affiliation{%
  \institution{The University of Alabama}
  \city{Tuscaloosa}
  \state{Alabama}
  \country{USA}}
\email{ryu5@ua.edu}

\renewcommand{\shortauthors}{Dai et al.}

\begin{abstract}
Partial differential equations (PDEs) govern nearly every physical process in science and engineering, but solving them at scale remains prohibitively expensive. Generative AI has transformed language, vision, and protein science, but learned PDE solvers have not undergone a comparable shift. Existing paradigms each capture part of the problem. Physics-informed neural networks embed residual structure, although they are often difficult to optimize in stiff, multiscale, or large-domain regimes. Neural operators amortize across instances, although they commonly inherit a snapshot-prediction view of solving and can degrade over long rollouts. Diffusion-based solvers model uncertainty, although they are often built on a solver template that still centers on state regression. We argue that the core issue is the abstraction used to train learned solvers. Many models are asked to predict states, while many scientific settings require modeling how uncertainty moves through constrained dynamics. The relevant object is transport over physically admissible futures. This motivates \emph{flow learners}: models that parameterize transport vector fields and generate trajectories through integration, echoing the continuous dynamics that define PDE evolution. This physics-to-physics alignment supports continuous-time prediction, native uncertainty quantification, and new opportunities for physics-aware solver design. We explain why transport-based learning offers a stronger organizing principle for learned PDE solving and outline the research agenda that follows from this shift.
\end{abstract}


\maketitle


\section{Introduction}

Consider the forecast problem seventy-two hours before a hurricane makes landfall. The goal is not to predict one atmospheric state a few hours ahead. The goal is to represent multiple physically possible futures: different tracks, wind fields, and storm surges that may unfold over the next several days under sparse observations and model uncertainty. A learned solver trained to predict the next state can look accurate in the short term and still fail in practice. If two plausible storm tracks diverge, their average may form an artificial track that no real storm would follow, even though evacuation and infrastructure decisions may be based on that prediction. The problem is that the model is learning the wrong object.

This tension is not unique to weather. Partial differential equations (PDEs) are the language of modern scientific computing~\cite{evans2010pde}. When a surgeon needs fast hemodynamic prediction through a prosthetic valve, the task is a Navier--Stokes solve. When an engineer certifies a turbine blade or battery pack under uncertain thermal and mechanical load, the safety case depends on large ensembles of PDE solves. When climate scientists assess risk under alternative forcing scenarios, they need many physically coherent futures, not a single forecast. The practical bottleneck is no longer whether PDEs matter. It is whether we can solve them fast enough, often enough, and with uncertainty that is useful for action.

That bottleneck remains severe. Direct numerical simulation of turbulent flow at realistic Reynolds numbers can require on the order of $10^{12}$ grid points and weeks on leadership-class systems~\cite{choi2012dns,dai2026pest}. Patient-specific cardiac simulation may take 12--48 hours, which is too slow for real-time clinical use. Cloud-resolving climate projection remains expensive even at exascale~\cite{schneider2017climate}. These costs do more than slow current workflows. They block whole classes of inverse design, adaptive control, planning, and rapid what-if analysis.

Machine learning should have changed this already. So far, it has not. Physics-informed neural networks (PINNs) encode residual structure, although they often become difficult to optimize in stiff, multiscale, or large-domain regimes~\cite{raissi2019pinn,krishnapriyan2021pinn_failure}. Neural operators amortize across instances, although their dominant use still treats solving as a map between snapshots, and long rollouts remain fragile~\cite{li2021fno,lu2021deeponet,brandstetter2022mp,lippe2024pderefiner,serrano2024aroma,ruhe2024rolling}. Diffusion-style PDE models recover uncertainty, although they often function as generative layers placed on top of a solver template that still centers on regression~\cite{huang2024diffusionpde,gencast2024}. The field has progressed, but its default mental model of what a learned solver should learn has barely changed.

This paper advances a sharper claim. In many scientifically important PDE regimes, learned solving has been organized around the wrong abstraction. The dominant template asks a model to predict states, such as $u_t \mapsto u_{t+\Delta t}$, or sometimes $u_0 \mapsto u_T$. That framing is easy to supervise. It is often a poor fit to the scientific object. In chaotic, partially observed, multiscale, or decision-relevant settings, the central object is transport over a set of physically admissible futures. A stronger target is therefore a transport law over structured state distributions.

We use the term \emph{flow learner} for models built around that abstraction. A flow learner parameterizes a vector field over physical states or latent physical representations and produces predictions by integrating or sampling the induced dynamics. The appeal of this view is not stylistic. A PDE is itself a law of continuous transport on a constrained state space, and flow learners are built around the same primitive. That is what we mean by \emph{physics-to-physics}. The phrase is literal. It names an alignment between solver structure and physical evolution.

The rest of the paper develops this claim. We first explain why the regression default is a poor fit for many hard PDE regimes. We then define flow learners more carefully, make the physics-to-physics view concrete, and describe the research agenda that emerges once transport becomes the default object of learning.

\section{The Regression Default}
\label{sec:regression}

The central claim of this paper begins with a simple observation: most learned PDE solvers are trained as state predictors. They map one state to the next, or one initial state to a later one, and treat solving as a supervised regression problem over snapshots. That formulation is natural, and it has supported real progress. It is also an increasingly poor fit to the scientific regimes that motivate learned solving in the first place. Once long horizons, uncertainty, partial observability, or decision relevance enter the picture, the regression view starts to miss the structure that the solver is actually supposed to capture.

A useful way to see this is to ask what a solver actually represents. For a deterministic PDE, the key object is a family of evolution operators, or semigroup, that carries admissible states forward in time. Under uncertainty, the object becomes richer: one also needs the induced transport of distributions over those states. A one-step regressor approximates only a small piece of that structure. In benign regimes, that surrogate may be sufficient. In long-horizon, chaotic, or partially observed regimes, it often is not.

\smallskip\noindent\textbf{A local map does not yet define a solver.}
Learning $u_t \mapsto u_{t+\Delta t}$ with low one-step error does not mean the model has learned the evolution law that matters after hundreds or thousands of compositions. Small biases accumulate into drift. Missing modes grow into instability. Approximation error pushes trajectories away from the physically admissible manifold. This is why long-rollout neural PDE models often need corrective refinement or auxiliary stabilization~\cite{brandstetter2022mp,lippe2024pderefiner,kohl2026benchmarking}. Classical integrators repeatedly consult the governing dynamics and adjust at intermediate steps. A learned model that only reproduces local transitions has captured outputs at short range. It has not yet captured the dynamics that define the solver.

\smallskip\noindent\textbf{Point prediction often misses the scientific target.}
When observations are sparse, initial conditions are uncertain, or the dynamics are chaotic, the target is a conditional distribution over futures. This is a practical requirement, not a philosophical preference. In the hurricane example, decision makers care about probability mass over landfall corridors, surge levels, and damaging winds. A model trained with pointwise loss can compress that structure into a conditional mean that does not correspond to any admissible storm. The same issue appears in turbulence, wildfire spread, hemodynamics under uncertain boundary conditions, and inverse design under uncertain forcing. In such settings, a single ``best'' field is often an artifact of the loss function rather than the scientific quantity of interest~\cite{huang2024diffusionpde,gencast2024,shysheya2024conditional}.

\smallskip\noindent\textbf{Adding physics after the fact leaves the core dynamics unconstrained.}
The regression default also changes where physical structure enters the pipeline. PINNs insert PDE residuals into the loss, but this often leads to brittle optimization in stiff, multiscale, or large-domain settings~\cite{raissi2019pinn,krishnapriyan2021pinn_failure,lu2024generative}. Data-driven operators amortize efficiently across instances, yet unless structure is built into the model, the learned map can still violate conservation, dissipation, symmetry, incompressibility, or realizability. Equivariance helps with some symmetries~\cite{satorras2021egnn,jacobsen2025cocogen}, but it does not by itself guarantee physically admissible evolution. Across much of the literature, physics enters as a regularizer, a projection step, or a repair mechanism applied after the main map has already been learned. That ordering is a poor fit for scientific computing.

These three problems share a common source. They reflect an abstraction mismatch. The field has largely optimized a convenient surrogate of the solver, not the scientific object the solver is supposed to represent. For many important PDE regimes, the default learned object should be continuous, distributional, and physics-native (i.e., structured to reflect the governing physical law rather than added after the fact). Table~\ref{tab:comparison} summarizes the central contrast of this paper. 

\begin{table}[h]
\caption{Native capabilities of learned PDE solver paradigms.}
\label{tab:comparison}
\centering
\small
\begin{tabular}{lccc}
\toprule
\textbf{Paradigm} & \shortstack{\textbf{Learns a}\\ \textbf{transport law}} & \shortstack{\textbf{Native future}\\ \textbf{distribution}} & \shortstack{\textbf{Physics inside}\\ \textbf{the dynamics}} \\
\midrule
PINNs & Limited & Limited & Partial \\
Neural operators & Limited & Limited & Limited \\
Diffusion solvers & Partial & Native & Limited \\
\textbf{Flow learners} & \textbf{Native} & \textbf{Native} & \textbf{Native} \\
\bottomrule
\end{tabular}
\end{table}



\begin{figure}[t]
\centering
\includegraphics[width=\columnwidth]{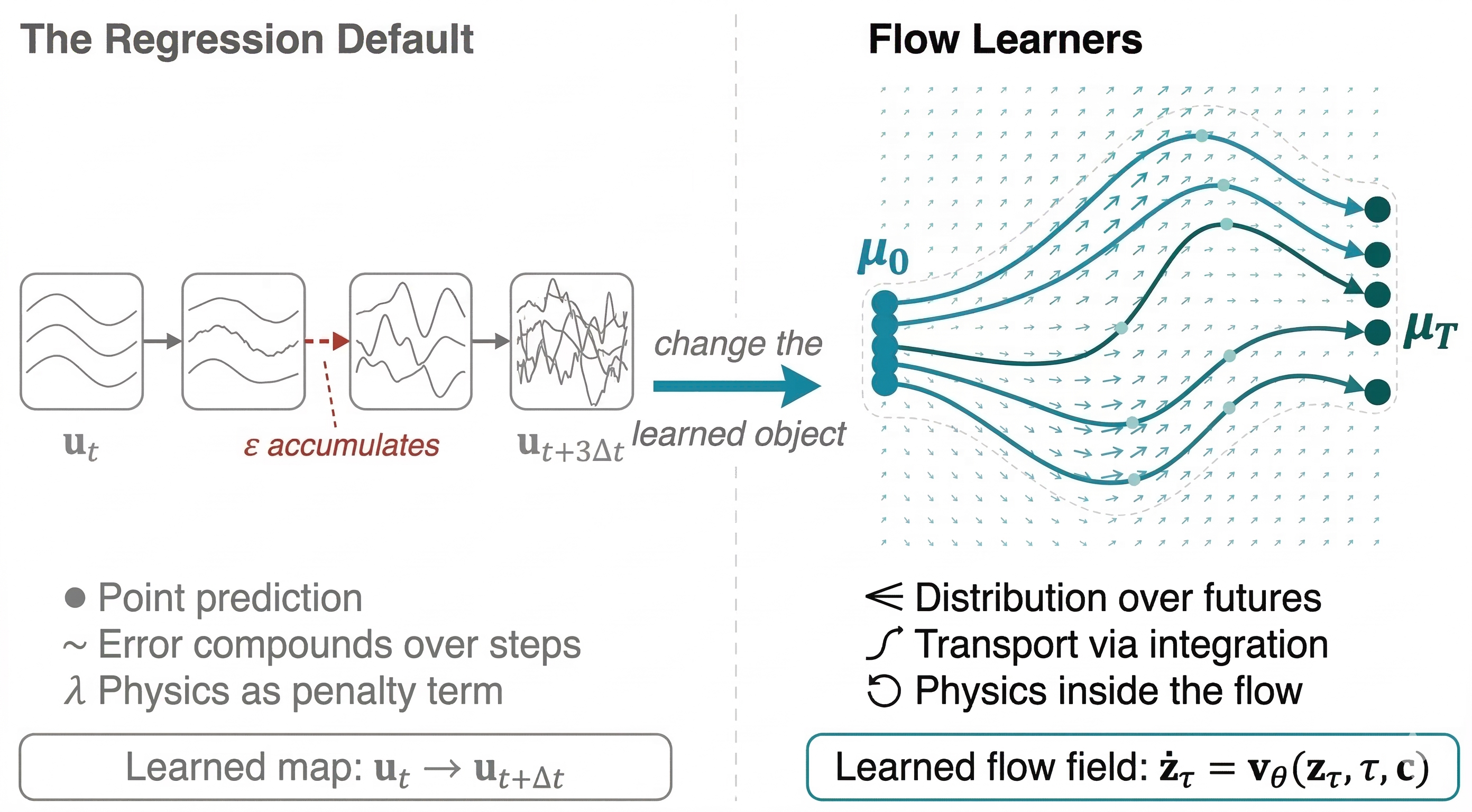}
\caption{A transport view of learned PDE solving. The solver learns a transport field whose integration generates physically admissible futures and ensembles.}
\label{fig:overview}
\end{figure}

\section{A Physics-to-Physics View}
\label{sec:idea}

\begin{figure}[t]
\centering
\includegraphics[width=\columnwidth]{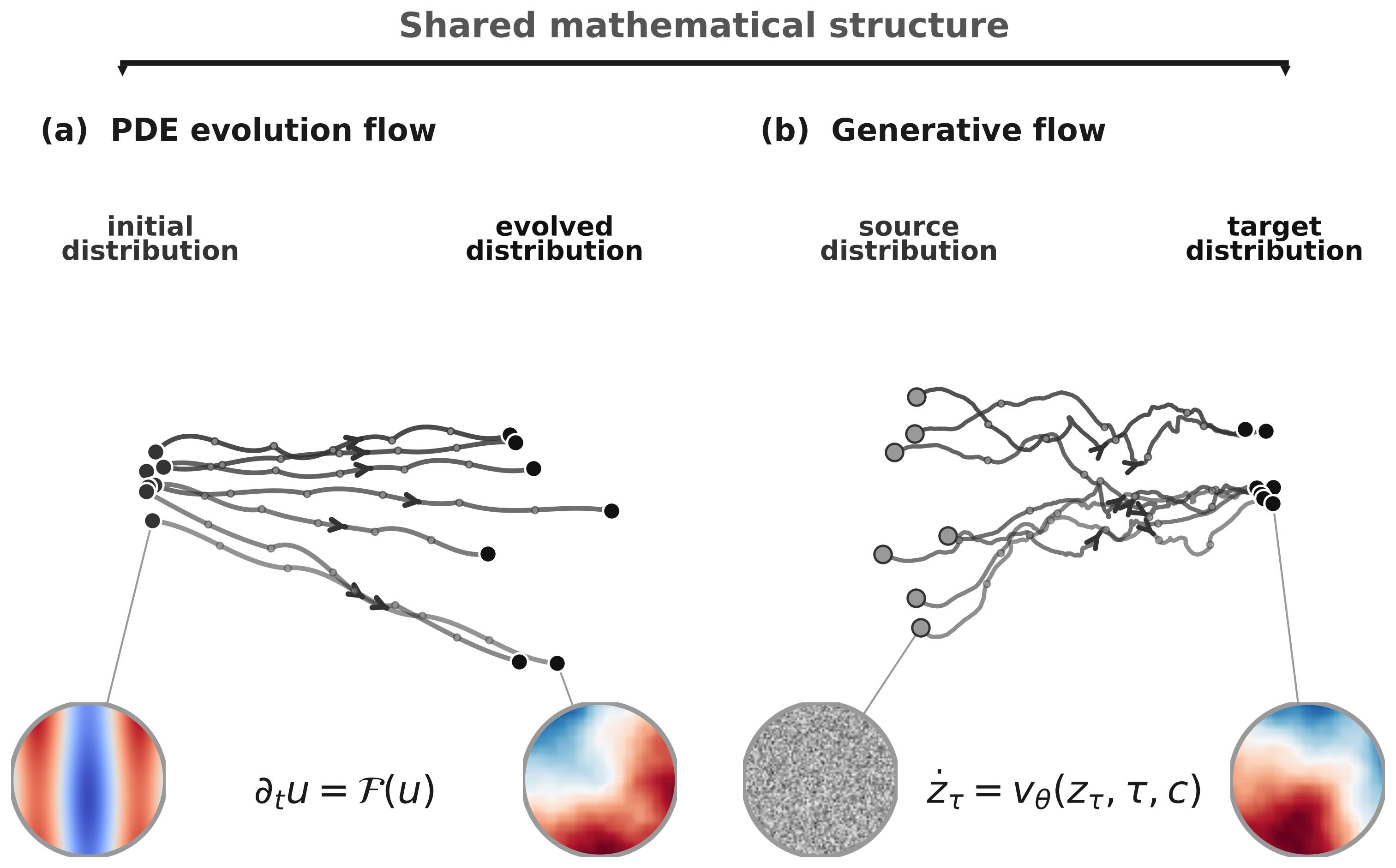}
\caption{Why ``physics-to-physics'' is literal. Both PDE evolution ($\partial_t u = \mathcal{F}(u)$) and flow learners generate trajectories by integrating a vector field; the differences lie in the state space, conditioning, and time parameterization.}
\label{fig:phase-space}
\end{figure}

Once the limits of the regression default are clear, the next question is what the solver should learn instead. Our answer is transport. The proposal in this section is therefore a change in solver abstraction, not a preference for one more generative architecture. Figures~\ref{fig:overview} and~\ref{fig:phase-space} illustrate this shift. The solver should learn transport as its primary object, because transport is the primitive that already organizes PDE evolution itself.

\subsection{The Scientific Object}

Let $\mathcal{M}$ denote the admissible state space induced by geometry, boundary conditions, constitutive structure, and hard physical constraints. A deterministic PDE defines an evolution operator
\[
S_t : \mathcal{M} \rightarrow \mathcal{M},
\]
which carries an admissible initial condition to a later admissible state. Once initial conditions, forcing, or parameters become uncertain---or once the system is only partially observed---the relevant initial object is no longer a single point in $\mathcal{M}$. It is a distribution $\mu_0$ over $\mathcal{M}$, often a posterior conditioned on observations. The dynamics then transport that distribution according to
\[
\mu_t = (S_t)_\# \mu_0.
\]

This equation sits at the center of our argument. In many practical scientific settings, users care more about $\mu_t$ than about any single realization. They care about tail risk, failure probabilities, credible intervals, return times, robust design margins, or ensembles of plausible future fields. Even when the governing PDE is deterministic, the \emph{solver task} becomes distributional as soon as the inputs are uncertain, the observations are sparse, or the downstream decision depends on rare but consequential events. The learned object should reflect that reality.

State-regression methods usually approximate one element of the family $\{S_t\}_{t \ge 0}$, often $S_{\Delta t}$, and then compose it. That surrogate can work when uncertainty is narrow, observations are dense, and horizons are short. It is a weak default for the harder regimes that motivate learned PDE solving in the first place. Flow learners target the transport law more directly. They model how trajectories move through admissible state space and, under uncertainty, how distributions of trajectories evolve.

The hurricane example makes this concrete: the forecast problem is to transport a posterior over atmospheric states into a distribution over physically plausible storm futures. The same logic applies to turbulence with uncertain inflow, cardiac flow with patient-specific boundaries, and climate response under uncertain forcing.

\subsection{Flow Learners as a Solver Abstraction}

We use \emph{flow learner} as a solver abstraction. The term is broader than any single algorithmic family. A model counts as a flow learner here if its primary learned object is a transport vector field over physical states or latent physical representations, and if prediction is produced by integrating or sampling that field. In its most generic form, the learned dynamics can be written as
\[
\dot z_{\tau} = v_{\theta}(z_{\tau}, \tau, c),
\]
where $z_{\tau}$ is a state or latent state, $\tau$ is integration time, and $c$ denotes conditioning information such as observations, geometry, forcing, boundary conditions, or coarse simulation context. When initialized from a source distribution, latent prior, or posterior over states, the dynamics transport an ensemble rather than a single point.

This definition focuses on the learned object. It separates the paradigm from questions of method branding. A diffusion or score model falls into this view only when it is used as a transport model over solution distributions, for example through reverse-time or probability-flow dynamics~\cite{song2021score}. A neural operator fits only when it parameterizes transport rather than a static state map. Neural ODEs are closely related; in their standard usage they integrate a learned vector field along a single deterministic trajectory, though they can be combined with stochastic initial conditions or latent priors to produce ensemble predictions. Flow matching is especially important because it offers a practical route to learning transport fields directly, though it is best understood as a training principle within the broader paradigm~\cite{lipman2023flow,lim2024elucidating,li2024fourier,zhou2025score,li2025generative}.

This taxonomy matters because it keeps the proposal from collapsing into a method survey. The paper does not champion one algorithmic family. The central claim is that the field should change its default object of learning, which many algorithms may instantiate. The key question is whether the solver learns transport.

\subsection{Why Physics-to-Physics}

The phrase \emph{physics-to-physics} can sound rhetorical unless it is defined carefully. We use it in a literal computational sense. PDE evolution and flow learners share four structural primitives.

First, both are organized around \emph{vector fields}. A PDE specifies tangent directions over states; a flow learner also learns tangent directions over states or latent physical representations. Second, both generate outcomes through \emph{integration}. A PDE solution emerges by integrating governing dynamics, and a flow learner generates a trajectory or sample by integrating its transport field. Third, both act on \emph{constrained state spaces}. Physical dynamics do not evolve over arbitrary tensors. They evolve over manifolds shaped by geometry, conservation laws, constitutive relations, and admissibility. A flow learner can be built on these same spaces, or on latent spaces designed to preserve them. Fourth, under uncertainty, both induce \emph{transport of measures}. Physical uncertainty is the pushforward of distributions under dynamics, and flow learners are designed to represent exactly that object.

This structural alignment matters because it changes where solver intelligence lives. Once the vector field becomes the main learned object, long-horizon prediction becomes integration of learned dynamics. Uncertainty becomes an evolving distribution produced by transport. Physical structure can be placed directly in the field parameterization, the admissible manifold, the transport path, or the sampling procedure itself~\cite{divfree2024,holderrieth2025pcfm,dong2026synergizing}.

Some caveats are important. Generative time and physical time need not coincide. Noise-to-data transport should not be confused with physical rollout. And not every score model is automatically a solver. Even with those caveats, the design principle remains strong: a transport-based model tracks the mathematical form of physical evolution more closely than a snapshot regressor does.

A natural objection is that continuous transport sounds elegant but difficult to train. That objection has weakened substantially. Classical Neural ODE training requires differentiation through ODE solves, which can be expensive and unstable~\cite{chen2018neural}. Flow matching and related objectives relax that burden by turning training into direct vector-field regression~\cite{lipman2023flow}. In PDE settings, derivative supervision can often be approximated from trajectories via finite differences, splines, or spectral estimates, and continuous-time flow operators already point toward arbitrary-time querying and improved long-horizon behavior~\cite{ai2025cfo,subhash2026splineflow}. The proposal is therefore no longer only conceptually appealing. It is becoming technically plausible.

\section{Why This Argument Is Timely Now}

The transport view would have sounded premature a few years ago, when continuous-time generative modeling for scientific dynamics was still technically fragile, and physically constrained generation was far less developed. It sounds much more plausible now because several ingredients have started to align.

\smallskip\noindent\textbf{Transport learning at scale has become practical.}
The flow-matching framework showed that transport vector fields can be learned without differentiating through long ODE trajectories during training~\cite{lipman2023flow}. In PDE settings, continuous-time operator variants already support arbitrary-time querying and improved long-horizon behavior from irregular trajectories~\cite{ai2025cfo,dai2026flowrefiner}. The computational objection has not vanished, but it is no longer decisive.

\smallskip\noindent\textbf{Generative scientific modeling has moved beyond speculation.}
Diffusion and related models have entered weather ensembles, partially observed PDE inference, and function-space generation~\cite{huang2024diffusionpde,gencast2024,ofm2024,lim2024fundiff,andrae2024continuous,ruhling2024probablistic,li2023seeds,chen2026earth}. The field no longer needs to ask whether generative models can represent scientific states. The more urgent question is whether generation will remain a wrapper around regression or become part of a transport-first solver design.

\smallskip\noindent\textbf{Constraint-aware transport is maturing.}
Stronger forms of physical enforcement during generation are becoming technically viable, which reduces exclusive reliance on fragile soft penalties~\cite{divfree2024,holderrieth2025pcfm,lipman2023flow}. This matters because scientific users will not trust samples that look plausible yet violate the laws that make them actionable.

\smallskip\noindent\textbf{The data regime has changed.}
Large simulation archives and early physics foundation models are making pretraining, retrieval, transfer, and active archive curation realistic research targets rather than distant aspirations~\cite{physix2025,shi2025flowmarching,chen2026latent}. From a KDD perspective, this is crucial. Once the solver is transport-based and distributional, the data strategy becomes part of the solver itself.

\section{A Transport-First Research Agenda}

Changing the learned object from state prediction to transport does more than suggest a different model family. It reorganizes the research agenda around a different set of priorities. Questions that once looked optional begin to move to the center, because they now follow directly from what the solver is supposed to represent. The point of this section is therefore to spell out the agenda that naturally emerges once learned PDE solving is organized around transport over physical futures.

\smallskip\noindent\textbf{Benchmark the law of motion.}
Once transport becomes the learned object, one-step RMSE should lose its status as the headline metric. The central empirical comparison is no longer between two local regressors with different backbones. It is between composing snapshot maps and integrating learned dynamics. That comparison should be judged by long-horizon skill at matched compute, robustness under irregular sampling, arbitrary-time querying, and preservation of physically meaningful structure over long rollout.

This shift in benchmarking changes what counts as progress. A solver that excels at $u_t \mapsto u_{t+\Delta t}$ yet drifts off the admissible manifold by step 200 is weak in the regimes that matter most. A model that cannot answer reverse-time, missing-time, or counterfactual-time queries without retraining remains tied to the discrete habits of state regression. If transport is the right abstraction, the field should benchmark semigroup quality rather than local fit alone~\cite{armegioiu2026memory}.

\smallskip\noindent\textbf{Treat uncertainty as a solver output.}
For chaotic and partially observed PDEs, ensemble prediction should become standard. Weather made this lesson unavoidable decades ago, yet learned PDE solving still often treats uncertainty as an optional add-on. A transport-first view removes that separation. The native solver output is a distribution over physically admissible futures~\cite{koupai2025enma}.

That shift also demands a different evaluation culture. Proper scoring rules, sharpness, coverage, rare-event skill, tail calibration, and physical admissibility should be reported together. A sharp model without calibration is dangerous. A calibrated model that breaks physical structure is equally dangerous. Under this agenda, generative evaluation is solver evaluation. This is one of the clearest ways in which learned PDE solving becomes a KDD problem rather than a narrow numerical-analysis exercise.

\smallskip\noindent\textbf{Make data acquisition part of the solver.}
When each label may cost hours or days of simulation, the data policy is inseparable from the solver design. A transport model can only represent the branches of the dynamics it has seen, or has been guided to query. Active learning should therefore determine which initial conditions, boundary regimes, geometries, parameters, resolutions, or fidelity levels are worth simulating next. Historical archives should be searched, deduplicated, reweighted, and scheduled with the same care currently devoted to architecture search.

This becomes even more important under partial observability. Rare but consequential branches of the dynamics are exactly the ones that simple point prediction tends to wash out and that decision makers care about most. If those branches are missing from the archive, a transport learner may still collapse them, only with more elaborate machinery. Under the new paradigm, a good solver is inseparable from a good data strategy. Archive curation, retrieval, curriculum design, and multi-fidelity transfer are all first-order components of the solver.

\smallskip\noindent\textbf{Put physics inside the loop.}
Scientific users will not trust models that are statistically plausible yet physically inadmissible. If transport is the learned object, physics should shape transport itself. This suggests a clear agenda: divergence-free or symmetry-preserving field parameterizations, constrained latent manifolds, projection-guided sampling, residual-aware correctors, transport paths inside admissible regions, and explicit characterization of the tradeoff among fidelity, calibration, and constraint satisfaction~\cite{divfree2024,holderrieth2025pcfm,millard2026particle}.

The location of the constraint is the key point. Penalizing final samples after an unconstrained generative process has unfolded is usually too late. By then, much of the model capacity has already been spent learning dynamics that should have been excluded from the start. The scientific goal is decision-grade reliability under a hard structure. That goal is far more achievable when constraints live inside the transport loop.

\section{Challenges and Milestones}

A proposal at the level of solver abstraction should ultimately be judged by what it takes to make it real. For the transport view, the central issue is no longer conceptual appeal alone. The harder question is whether transport over physical futures can be turned into a solver that remains trainable, auditable, and useful in the regimes that motivated the proposal in the first place: branching futures, sparse observations, extreme dimensionality, irregular geometry, and decision settings where calibration and physical admissibility matter as much as point accuracy. That is where the true challenges of the agenda appear.

\smallskip\noindent\textbf{Coverage of branching futures.}
Flow learners are most useful when futures branch under uncertainty. The same setting also makes them data-hungry in a specific way: the archive should cover the rare, heterogeneous, or tail events that matter for decisions. If the data populate only common regimes, a transport model may still collapse uncertainty, only more smoothly than a point regressor. The issue is not sample count alone. It is support for the branches of the dynamics that users actually care about.

\smallskip\noindent\textbf{Scaling to extreme dimension and geometry.}
Real PDE systems involve millions of degrees of freedom, irregular meshes, moving boundaries, multi-physics coupling, and mixed scales. Transport on such spaces will require latent representations, adaptive discretization, geometry-aware parameterization, and efficient integrators. These are central scientific questions for the agenda.

\smallskip\noindent\textbf{Verification and trust.}
A generated field that looks plausible is not evidence of correctness. Engineers and scientists need residual audits, conservation checks, constraint-satisfaction rates, calibration diagnostics, and eventually error characterizations strong enough to support downstream choices. If transport-based solvers cannot be audited in ways that scientific users trust, the conceptual gain will not translate into adoption.

\smallskip\noindent\textbf{Inference cost.}
Integrating a transport field is often slower than a single forward pass through a deterministic operator. Straight transports, distilled samplers, amortized correctors, and few-step generation matter for this reason. Without such advances, the field may end up with a compelling abstraction that is still too slow for the operational settings where surrogate solvers are most valuable.

We would view the paradigm as succeeding over the next five years if the community can meet several falsifiable milestones:
\begin{itemize}
    \item \textbf{Matched-compute long-horizon wins.} On at least one established chaotic or multiscale PDE benchmark, transport-based solvers outperform strong next-state regressors on long-horizon \emph{distributional} skill, not one-step error.

    \item \textbf{Joint reporting of accuracy, calibration, and physics.} Generative PDE papers routinely report proper scoring rules together with conservation error, residual audits, and constraint violation rates, rather than one of these in isolation.

    \item \textbf{Decision-grade ensemble value.} In a representative application class, such as weather risk, hemodynamic planning, or CFD under uncertain inflow, transport-based ensembles improve downstream decisions at useful latency and cost.

    \item \textbf{Order-of-magnitude label savings.} Active acquisition or archive-aware training reduces the number of expensive simulations needed to reach a target long-horizon skill by roughly an order of magnitude on a nontrivial PDE family.

    \item \textbf{Cross-PDE transport pretraining.} Pretrained flow learner transfers across multiple PDE families, geometries, or resolutions with modest adaptation data while preserving both calibration and physical admissibility.
\end{itemize}

Evidence against the thesis would be just as informative. If transport-based solvers continue to lose to strong autoregressive baselines on the metrics that matter after matched compute, or if constraint satisfaction and calibration remain too brittle, then the field should revise the claim. A Blue Sky argument earns its place by being sharp enough to fail.

Success, however, does not require replacing numerical solvers. Classical solvers will remain the source of labels, verification, and, in many settings, final authority. Success means something more specific: learned transport becomes the preferred amortized layer for ensemble generation, rapid exploration, inverse design, uncertainty-aware forecasting, and decision support in regimes where repeated full solves are too expensive.



\begin{acks}
The authors acknowledge partial support from the National Science Foundation (NSF) under grants 2203581, 2239175, 2316305, 2147195, 2425845, and 2530609; the USGS award G22AC00266; the NASA grants 80NSSC24K1061 and 80NSSC25K0013; NSF NCAR's Derecho HPC system; and the Alabama Center for the Advancement of Artificial Intelligence Faculty Fellowship Program.
\end{acks}

\bibliographystyle{ACM-Reference-Format}
\bibliography{references}

\end{document}